\documentclass[runningheads]{llncs}
\usepackage{esvect}
\usepackage[T1]{fontenc}
\usepackage{caption}
\usepackage{graphicx,verbatim}
\usepackage{amsmath}
\usepackage{amssymb}   
\usepackage{amsfonts}
\usepackage{bm}       
\usepackage{booktabs}  
\usepackage{graphicx}  
\usepackage{booktabs}
\usepackage{multirow}
\usepackage{graphicx}
\usepackage[table]{xcolor}
\usepackage{multirow}
\usepackage{tikz}
\usepackage{pgfplots}
\usepackage{amsmath}
\DeclareMathOperator*{\argmax}{arg\,max}
\pgfplotsset{compat=1.17} 
\usepackage{booktabs}
\usepackage{array}
\usepackage{marvosym}
\DeclareRobustCommand{\corr}{\textsuperscript{\scriptsize\Letter}}

\usepackage{xcolor}
\usepackage[
  colorlinks=true,
  linkcolor=blue,
  citecolor=green,
  urlcolor=blue,
  filecolor=blue
]{hyperref}

\newcommand{\sj}[1]{{\color{red}{[SJ:#1]}}}

\begin{document}
\title{Evo-RAD: Navigating Rare Retinal Disease Diagnosis via Self-Evolving Agentic Retrieval}
% \title{GRPO-Active Retrieval Adapter: Navigating the Long-tail of Retinal Disease Diagnosis via Iterative Set Refinement }
\titlerunning{Evo-RAD: Self-Evolving Agentic Retrieval for Rare Retinal Diagnosis
}

\author{Wangding Xia\inst{1}*
\and
Ye Du\inst{1}*
\and
Jiashi Lin\inst{2}
\and
Meng Wang\inst{3,4}
\and
Danli Shi\inst{5,6}
\and
Shujun Wang\inst{1}\corr
}
\authorrunning{W. Xia et al.}
\institute{Department of Biomedical Engineering, The Hong Kong Polytechnic University, Hong Kong
SAR, China
\and School of Computer Science, Northwestern Polytechnical University, Xi'an, China
\and Centre for Innovation and Precision Eye Health, Yong Loo Lin School of Medicine, National University of Singapore, Singapore 119228, Singapore
\and Department of Ophthalmology, Yong Loo Lin School of Medicine, National University of Singapore, Singapore 119228, Singapore
\and School of Optometry, The Hong Kong Polytechnic University, Hong Kong SAR, China
\and Research Centre for SHARP Vision (RCSV), The Hong Kong Polytechnic University, Hong Kong SAR, China \\
\email{shu-jun.wang@polyu.edu.hk}\\
\Letter~Corresponding author\\
*These authors contributed equally to this work.
}
  
\maketitle              % typeset the header of the contribution
\begin{abstract}

Large-scale pretrained foundation models have revolutionized general medical screening, but often falter on rare diseases because such conditions are underrepresented in real-world clinical datasets.
While retrieval-augmented diagnosis attempts to mitigate this, conventional static methods frequently succumb to the \textit{hubness problem}, retrieving visually similar but semantically incorrect common diseases. To address this, we propose \textbf{Evo-RAD}, a self-evolving agentic framework that transforms evidence acquisition into a dynamic decision-making task.
We formulate retrieval as a \textit{Markov Decision Process} (MDP) where a graph-based agent observes the reference set \textit{state} and executes \textit{actions} to purge discordant evidence (DELETE), acquire pathologically consistent samples (INSERT), or conclude the evolution (TERMINATE). Optimized via Group Relative Policy Optimization (GRPO) with a homogeneity-aware reward, the agent learns to maximize the diagnostic homogeneity of the support reference set.  Experiments on retinal disease benchmarks show that Evo-RAD substantially improves rare-disease diagnosis, outperforming retinal foundation models by \textbf{+21.04\%}, while also surpassing retrieval-based and parameter-efficient fine-tuning methods by \textbf{+3.56\%}. Code is available at \href{https://github.com/SDH-Lab/Evo-RAD}{https://github.com/SDH-Lab/Evo-RAD}.


% \keywords{Long-tailed Diagnosis  \and Rare Disease Adaptation \and Group Relative Policy Optimization \and Active Retrieval \and Foundation Model Adapter.}
% Authors must provide keywords and are not allowed to remove this Keyword section.

\end{abstract}

\section{Introduction}
% Foundation models are rapidly reshaping ophthalmic diagnosis by learning unified representations across retinal images and clinical language~\cite{zhang2025biomedclip,Wang2022Medclip,Shi2025Vision,Silva2025Foundation,Wang2025Enhancing}. Large-scale Vision-Language Models (VLMs) such as VisionFM~\cite{Qiu2024Development} and RetiZero \cite{Wang2025Enhancing} achieve strong benchmark results by aligning retinal imagery with clinical narratives, making them attractive backbones for automated screening. However, rare retinal diseases remain challenging: they are infrequent and clinically heterogeneous~\cite{phillips2024time}, so their features are weakly constrained during pretraining and can be entangled with visually similar common conditions, leading to degraded real-world performance (lower sensitivity and higher confusion with prevalent diagnoses).

Foundation models are rapidly reshaping ophthalmic diagnosis by learning unified representations across retinal images and clinical language~\cite{zhang2025biomedclip,Wang2022Medclip,Shi2025Vision,Silva2025Foundation,Wang2025Enhancing}. Recent Vision-Language Models (VLMs), such as FLAIR~\cite{Silva2025Foundation} and RetiZero~\cite{Wang2025Enhancing}, demonstrate strong performance on public benchmarks by aligning retinal imagery with clinical narratives, and have thus become attractive for automated screening. However, deploying these VLMs into rare retinal diseases diagnosis remains challenging.
Rare diseases are infrequent and clinically heterogeneous~\cite{phillips2024time}, so their features are weakly constrained during pretraining and can be entangled with visually similar common diseases, leading to degraded real-world performance (lower sensitivity and higher confusion with prevalent diagnoses).
% are infrequently observed and clinically heterogeneous~\cite{phillips2024time}, so their representations are only weakly constrained during pretraining and are easily entangled with visually proximate but more prevalent conditions. Consequently, under real-world clinical distributions, VLMs often suffer substantial performance deterioration on rare diseases, typically reflected by reduced sensitivity and increased confusion with common diagnoses.

\if 0
Foundation models are rapidly reshaping ophthalmic diagnosis by enabling unified representations across images and clinical language~\cite{zhang2025biomedclip,Wang2022Medclip,Shi2025Vision,Silva2025Foundation,Wang2025Enhancing}. Recent large-scale vision–language models (VLMs), such as VisionFM~\cite{Qiu2024Development} and RetiZero~\cite{Wang2025Enhancing}, have achieved strong benchmark performance by aligning retinal imagery with clinical text at scale, making them attractive backbones for automated screening. Despite this impressive generality, deploying these models in real clinical settings remains challenging for rare retinal diseases. Rare pathologies are sparsely observed and highly heterogeneous, making their representations weakly constrained during pretraining and can be easily confounded with more prevalent conditions. As a result, when confronted with real-world clinical distributions, these models often exhibit pronounced degradation on rare retinal diseases.
\fi

To mitigate this degradation, an intuitive strategy is to adapt foundation models via \textit{Parameter-Efficient Fine-Tuning (PEFT)}, such as prompt tuning \cite{Zhou2022Learning,Zhan2024Xcoop,Koleilat2025Biomedcoop,du2025medical,Li2025DPC,luo2025llm} and adapter-based methods~\cite{Zhang2022Tip,Gao2021CLIP}. PEFT freezes the backbone and updates only a small set of additional or selected parameters, thereby reducing computational costs. However, PEFT can be unreliable for rare classes. With only a handful of rare cases, even this limited trainable capacity can overfit to acquisition-specific noise and spurious cues. Conversely, when trained on the full imbalanced corpus, optimization is dominated by common classes, biasing the learned adaptation toward prevalent conditions and limiting gains in rare-disease sensitivity where robustness is most critical.
% However, PEFT can be unreliable for rare classes. When only a handful of rare cases are available, even limited trainable parameters may overfit to acquisition-specific noise. When training on full imbalanced corpus, optimization is dominated by common classes, which can shift representations and limit performance where rare-disease sensitivity is most critical.

Besides PEFT, a more robust alternative is \textit{Retrieval-Augmented Diagnosis (RAD)}. Given a query image, RAD retrieves a small set of similar reference cases from an external database and conditions diagnosis on the retrieved examples as supporting evidence. However, most existing RAD pipelines~\cite{Shi2025Vision,Long2023Retrieval} still rely on one-shot nearest-neighbor retrieval with a fixed similarity metric, producing a static reference set. This rigidity design is fragile for rare diseases due to the \textit{hubness problem}~\cite{Radovanovic2010Hubs}, where a small subset of common-disease samples repeatedly appears as nearest neighbors. As a result, the retrieved evidence can be visually similar but semantically incorrect, which biases prediction toward common classes and undermines the reliability of retrieval-based support.

% To overcome the above limitations, we posit that retrieval for rare-disease diagnosis should be elevated from a passive matching operation to an iterative, context-aware reasoning process. In clinical practice, experts refine their working hypotheses by accumulating evidence, excluding mimicking conditions, and seeking patterns that remain consistent under scrutiny~\cite{croskerry2009universal}. Motivated by this, we propose \textbf{Evo-RAD}, a self-evolving agentic retrieval framework that upgrades evidence acquisition from \textit{passive feature matching} to \textit{dynamic decision making}. We cast retrieval as a \textit{Markov Decision Process (MDP)}~\cite{Sutton1998Reinforcement}, where a graph-based agent observes the \textit{state} of the current reference set and sequentially executes discrete \textit{actions} to refine it: purging discordant ``hub'' samples (\textbf{DELETE}), acquiring pathologically consistent samples from the external memory (\textbf{INSERT}), or concluding the evolution (\textbf{TERMINATE}). To effectively navigate this discrete action space, the agent is optimized via Group Relative Policy Optimization (GRPO)~\cite{guo2025deepseek} with a novel \textit{homogeneity-aware reward} that encourages a diagnostically coherent reference set. Throughout this mechanism, the final prediction is conditioned on progressively evolved evidence, enabling more faithful support for rare retinal disease diagnosis.

To overcome these limitations, we argue that retrieval for rare-disease diagnosis should move beyond passive nearest-neighbor matching to an iterative, context-aware reasoning process, analogous to how clinicians refine hypotheses by accumulating consistent evidence and excluding mimics~\cite{croskerry2009universal}. Motivated by this, we propose \textbf{Evo-RAD}, a self-evolving agentic retrieval framework that formulates retrieval as a \textit{Markov Decision Process (MDP)}~\cite{Sutton1998Reinforcement}: a graph-based agent observes the current reference-set \textit{state} and sequentially evolves it via \textbf{DELETE} (purge discordant hub samples), \textbf{INSERT} (add pathologically consistent cases from external memory), or \textbf{TERMINATE}. We optimize the policy with Group Relative Policy Optimization  (GRPO)~\cite{guo2025deepseek} under a \textit{homogeneity-aware reward} that promotes diagnostic coherence, so final prediction is conditioned on progressively evolved evidence, improving support for rare retinal disease diagnosis.

To summarize, our contributions are threefold. \textbf{First}, we propose \textbf{Evo-RAD}, an agentic retrieval framework moving beyond static similarity search to iterative, dynamic reference set refinement. \textbf{Second}, we design a \textit{homogeneity-aware} optimization objective and train the retrieval agent with GRPO, enabling discrete evidence evolving operations. \textbf{Third}, We conduct extensive experiments on retinal disease benchmarks, demonstrating that Evo-RAD yields a \textbf{+21.04\%} improvement over retinal foundation models and a \textbf{+3.56\%} gain over state-of-the-art static retrieval and PEFT methods.

\begin{figure}[t]
  \centering
  \includegraphics[width=\linewidth]{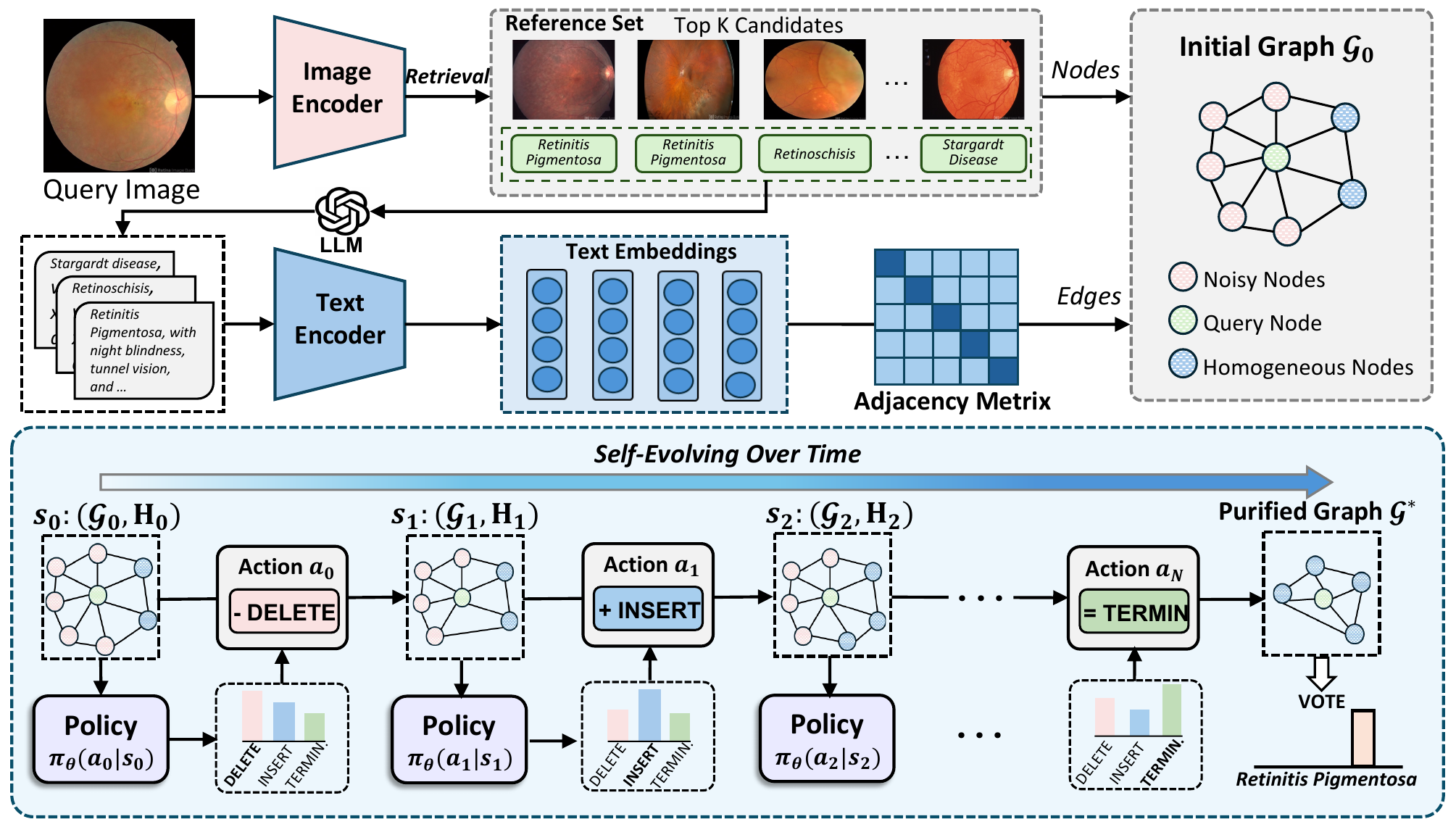}
  % %\vspace{-1em}
  \caption{\textbf{Overview of Evo-RAD.} Evo-RAD models retrieval as a Markov Decision Process where a graph policy evolves the evidence set through discrete \textsc{DELETE}, \textsc{INSERT}, and \textsc{TERMINATE} operations.}
  % \sj{insert a brief introduction.}} }
  % Conventional RAD performs one-shot nearest-neighbor retrieval under a fixed similarity metric, which is vulnerable to the \textit{hubness problem} in long-tailed embedding spaces. 
  % Evo-RAD reformulates retrieval as a self-evolving process: a graph-based agent observes the current reference set (\textit{state}) and iteratively refines it via discrete actions: \textsc{DELETE} discordant evidence, \textsc{INSERT} pathologically consistent candidates, or \textsc{TERMINATE} when evidence becomes sufficiently reliable.}
  %\vspace{-1.5em}
  \label{fig:evo-rad_overview}
\end{figure}

\section{Method}
Evo-RAD (Fig.~\ref{fig:evo-rad_overview}) is a self-evolving agentic retrieval framework for rare-disease diagnosis. 
It reformulates Top-$K$ retrieval as a sequential decision process that iteratively evolves the reference set to remove hub-driven distractors and admit clinically concordant cases, terminating once the evidence is diagnostically coherent. 
Technically, Evo-RAD introduces: (i) a Markov Decision Process-based self-evolving agentic retrieval formulation; (ii) a GRPO-trained graph policy that jointly optimizes query affinity and intra-set agreement; and (iii) a final evidence-conditioned prediction head.

\subsection{Markov Decision Process based Self-Evolving Agentic Retrieval}

Given a query image \(q\), we extract its visual embedding \(v_q\) using a frozen retinal foundation model and retrieve the Top-\(K\) most similar images with corresponding labels to form the initial reference set \(\mathcal{X}_0\subset\mathcal{D}\), where $\mathcal{D}$ is a rare disease database (\textit{e.g.}, the training subset).
To enable reference set update, we model the self-evolving agentic retrieval as a Markov Decision Process
$\mathcal{M}=\langle \mathcal{S},\mathcal{A},\mathcal{T},\mathcal{R}\rangle$~\cite{Sutton1998Reinforcement},
where the environment maintains an continual updating reference set \(\mathcal{X}_t\subset\mathcal{D}\) and the agent learns to evolve from \(\mathcal{X}_0\) to \(\mathcal{X}_t\) to suppress hub-induced distractors and retain diagnostically consistent references.

\vspace{3pt}

\noindent \textbf{\textit{1. State $\mathcal{S}$.}}
At time step $t$, the state $s_t \in \mathcal{S}$ summarizes the current reference set $\mathcal{X}_t$ with respect to the query $q$ and the intra-set semantic coherence.
We represent $s_t$ as a graph $s_t=(\mathcal{G}_t,\mathbf{H}_t)$, where $\mathcal{G}_t$ specifies the topology over the current references and $\mathbf{H}_t$ stores node features.

\textit{\textbf{1) Graph topology.}}
The topology $\mathcal{G}_t$ is instantiated by a semantic-consistency adjacency matrix defined within $\mathcal{X}_t$.
Because raw disease labels are often coarse, for each disease label $y\in\mathcal{Y}$ we perform clinical concept expansion using a large language model (GPT-5.1), producing a set of fine-grained clinical tags $\mathrm{Tags}(y)$ (e.g., ``bone-spicule pigmentation'' and ``cotton-wool spots'').
For each reference case $x$, we attach its tag text $\mathrm{Tags}(y_x)$ and encode it with the frozen text encoder of the vision--language model to obtain a semantic embedding $u_x$.
We then build the adjacency matrix $\mathbf{A}_t \in \mathbb{R}^{|\mathcal{X}_t|\times|\mathcal{X}_t|}$ by pairwise cosine similarity:
\begin{equation}
(\mathbf{A}_t)_{ij}
:= \frac{u_{x_i}^\top u_{x_j}}{\|u_{x_i}\|_2\,\|u_{x_j}\|_2},
\quad x_i,x_j\in\mathcal{X}_t.
\end{equation}
This construction links reference cases that share consistent expanded clinical semantics.

\textit{\textbf{2) Node features.}} The matrix $\mathbf{H}_t \in \mathbb{R}^{|\mathcal{X}_t| \times D}$ stacks node features $h(x) \in \mathbf{H}_t$, defined as:
\begin{equation}
h(x) = [\, v_x \| \phi_{\text{stat}}(x, \mathcal{X}_t, q) \,],
\end{equation}
where $\|$ denotes concatenation. Descriptor $\phi_{\text{stat}} \in \mathbb{R}^8$ stacks four base metrics and their mean deviations, capturing similarity between $x$ and: (i) query $q$ (visual), (ii) the initial rank, and set $\mathcal{X}_t$ regarding (iii) clinical text and (iv) disease categories.

\vspace{3pt}

\if 0
\noindent \textit{\textbf{2.Action $\mathcal{A}$.}}
We define a discrete action space
$\mathcal{A}=\{\mathcal{A}_{\text{del}}\cup\mathcal{A}_{\text{ins}} \cup a_{\text{term}}\}$. For deletion, an action \(a\in\mathcal{A}_{\text{del}}\) for each reference case $x \in \mathcal{X}_t$ removes it from the reference set, subject to minimum-size constraint \(|\mathcal{X}_t|\ge \textit{min\_size}\). For insertion, an action \(a\in\mathcal{A}_{\text{ins}}\) triggers adding a new reference from a pre-fetched candidate buffer \(\mathcal{B}\subset\mathcal{D}\). To avoid searching the full memory at each step, the environment deterministically inserts the most query-similar candidate $x_{\text{ins}}
=\argmax_{x\in\mathcal{B}\setminus \mathcal{X}_t}\text{sim}(q, x)$.
This design decouples \emph{when to expand} (learned by the agent) from \emph{what to insert} (selected by a visual prior), significantly reducing the effective action dimension. The agent terminates refinement with action \(a_{\text{terminate}}\) once \(\mathcal{X}_t\) is sufficiently coherent.
\fi 

\noindent \textit{\textbf{2. Action $\mathcal{A}$.}}
We use a discrete action space
$
\mathcal{A}=\mathcal{A}_{\text{del}} \cup \mathcal{A}_{\text{ins}} \cup \{a_{\text{term}}\}.
$
\textit{Deletion.} For each reference case $x\in\mathcal{X}_t$, a deletion action $a_{\text{del}}(x)\in\mathcal{A}_{\text{del}}$ removes $x$ from $\mathcal{X}_t$, subject to a minimum-size constraint $|\mathcal{X}_t|\ge \textit{min\_size}$.
\textit{Insertion.} An insertion action $a_{\text{ins}}\in\mathcal{A}_{\text{ins}}$ expands the set by adding one case from a pre-fetched candidate buffer $\mathcal{B}\subset\mathcal{D}$.
To avoid searching the full datastore at each step, the environment deterministically inserts the most query-similar candidate not already selected:
$
x_{\text{ins}}=\arg\max_{x\in \mathcal{B}\setminus \mathcal{X}_t}\mathrm{sim}(q,x).
$
This design lets the agent learn \emph{when} to expand the evidence set, while a fixed similarity prior determines \emph{what} to insert, thereby reducing the effective action dimensionality.
\textit{Termination.} The agent stops the refinement process by selecting $a_{\text{term}}$ when $\mathcal{X}_t$ is deemed sufficiently coherent.

\vspace{3pt}

\noindent \textbf{\textit{3. Transition $\mathcal{T}$.}}
Given action \(a_t\), transition \(\mathcal{T}\) updates the reference set as
\begin{equation}
\mathcal{X}_{t+1}=
\begin{cases}
\mathcal{X}_t\setminus\{x\}, & a_t\in\mathcal{A}_{\text{del}},\\
\mathcal{X}_t\cup\{x_{ins}\}, & a_t\in\mathcal{A}_{\text{ins}},\\
\mathcal{X}_t, & a_t=a_{\text{term}},
\end{cases}
\end{equation}
and reconstructs the graph \(\mathcal{G}_{t+1}\) and node features \(\mathbf{H}_{t+1}\) accordingly.

\vspace{3pt}

\noindent\textbf{\textit{4. Homogeneity-aware Reward \(\mathcal{R}\).}}
We design a \emph{homogeneity-aware reward} to encourage the agent to evolve a diagnostically coherent reference set. The total return of a trajectory \(\tau=\{(s_t,a_t)\}_{t=0}^{T}\) is
\begin{equation}
R(\tau)= R_{\text{traj}}(\mathcal{X}_T) + \sum_{t=0}^{T-1} r_{\text{step}}(s_t,a_t),
\end{equation}
where \(R_{\text{traj}}\) evaluates the quality of the final reference set and \(r_{\text{step}}\) provides immediate feedback for each evolving operation.

\textit{\textbf{1) Trajectory-Level Reward.}}
Let \(\hat{y}(\mathcal{X})\) be the final diagnosis conditioned on reference set \(\mathcal{X}\). To guide the policy toward label consistency and semantic homogeneity, we define the label purity \(\text{Pur}(\mathcal{X})\) and the mean pairwise text-density \(\text{Den}(\mathcal{X})\) as:
\begin{equation}
\text{Pur}(\mathcal{X}) = \frac{1}{|\mathcal{X}|}\sum_{x\in\mathcal{X}}\mathbb{I}[y_x=y^\star], \ \ 
\text{Den}(\mathcal{X}) = \frac{2}{|\mathcal{X}|(|\mathcal{X}|-1)}\sum_{i<j}\cos(u_{x_i},u_{x_j}),
\end{equation}
where \(y^\star\) is the ground-truth label and \(u_x\) is the text embedding of clinical tags for \(x\).
We then construct the trajectory reward as:
\begin{equation}
R_{\text{traj}}(\mathcal{X}_T)
=
\underbrace{\alpha\,\mathbb{I}\!\left[\hat{y}(\mathcal{X}_T)=y^\star\right]}_{R_{\text{acc}}}
+\underbrace{\beta\Big(\text{Pur}(\mathcal{X}_T)-\text{Pur}(\mathcal{X}_0)\Big)}_{R_{\text{purity}}}
+\underbrace{\gamma\,\text{Den}(\mathcal{X}_T)}_{R_{\text{density}}},
\end{equation}
which rewards correct prediction, purity gain over the initial set \(\mathcal{X}_0\), and intra-set semantic density.

% Let \(\hat{y}(\mathcal{X})\) be the final diagnosis conditioned on reference set \(\mathcal{X}\), and define the label purity
% \begin{equation}
% \text{Pur}(\mathcal{X})=\frac{1}{|\mathcal{X}|}\sum_{x\in\mathcal{X}}\mathbb{I}[y_x=y^\star],
% \end{equation}
% where \(y^\star\) is the ground-truth label of the query. To capture clinical semantic homogeneity, we define the mean pairwise text-density within \(\mathcal{X}\):
% \begin{equation}
% \text{Den}(\mathcal{X})=\frac{2}{|\mathcal{X}|(|\mathcal{X}|-1)}\sum_{i<j}\cos(u_{x_i},u_{x_j}),
% \end{equation}
% with \(u_x\) the text-encoder embedding of the expanded clinical tags attached to \(x\).
% We then set
% \begin{equation}
% R_{\text{traj}}(\mathcal{X}_T)
% =
% \underbrace{\alpha\,\mathbb{I}\!\left[\hat{y}(\mathcal{X}_T)=y^\star\right]}_{R_{\text{acc}}}
% +\underbrace{\beta\Big(\text{Pur}(\mathcal{X}_T)-\text{Pur}(\mathcal{X}_0)\Big)}_{R_{\text{purity}}}
% +\underbrace{\gamma\,\text{Den}(\mathcal{X}_T)}_{R_{\text{density}}},
% %\vspace{-0.8em}
% \end{equation}
% which rewards correct final prediction, purity improvement over the initial retrieval \(\mathcal{X}_0\), and intra-set semantic density.

\textit{\textbf{2) Step-Level Reward.}}
For each action, we provide positive-only feedback:
\begin{equation}
r_{\text{step}}(s_t,a_t)=
\underbrace{\eta_{\text{ins}}\;\mathbb{I}\!\left[a_t\in\mathcal{A}_{\text{ins}}\right]\mathbb{I}\!\left[y_{x_{\text{ins}}}=y^\star\right]}_{r_{\text{ins}}}
+
\underbrace{\eta_{\text{del}}\;\mathbb{I}\!\left[a_t\in\mathcal{A}_{\text{del}}\right]\mathbb{I}\!\left[y_{x_{\text{del}}}\neq y^\star\right]}_{r_{\text{del}}}.
%\vspace{-0.7em}
\end{equation}
We adopt a zero-penalty principle by assigning \(0\) reward to incorrect insertions/deletions, avoiding negative shaping that can hinder early exploration.

\if 0
To provide comprehensive supervision for the self-evolving process, we design a hierarchical reward system consisting of three trajectory-level components and two step-level components, all optimized under a zero-penalty principle.

The \textbf{trajectory-level reward} $R_{\text{traj}}$ evaluates the final reference set $X_T$ to ensure outcome quality:
\begin{equation}
R_{\text{traj}} = R_{\text{acc}} + R_{\text{purity}} + \lambda R_{\text{density}},
\end{equation}
where $R_{\text{acc}}$ combines base diagnostic accuracy with a resurrection bonus for correcting an initial failure; $R_{\text{purity}}$ measures both the absolute proportion of relevant cases and the relative purity gain over $X_0$; and $R_{\text{density}}$ acts as a semantic regularizer maximizing the intra-set clinical text density.

To guide the agent's exploration during training, we supplement the outcome with two \textbf{step-level rewards} $R_{\text{step}}$ that provide immediate feedback for each evolving action:
\begin{equation}
R_{\text{step}} = r_{\text{ins}} + r_{\text{del}},
\end{equation}
where $r_{\text{ins}}$ grants a $+3.0$ bonus for inserting a diagnostically relevant reference, and $r_{\text{del}}$ provides a $+1.0$ incentive for pruning a distractor. Crucially, we omit all negative penalties for incorrect actions to encourage bold exploration in early training stages, ensuring the policy focuses on learning positive clinical reasoning patterns.
\fi

\subsection{Relational Policy Learning via Group-Relative Optimization}
With the MDP defined above, we learn a retrieval evolving policy \(\pi_\theta(a_t \mid s_t)\) that iteratively updates the reference set \(\mathcal{X}_t\) toward a diagnostically consistent and semantically coherent subset.  Instead of treating instances independently, we instantiate \(\pi_\theta\) as a relational reasoner over the evidence graph and optimize it using a group-relative objective to stabilize learning without a value network.

\noindent \textbf{\textit{Graph Policy Network.}}
Given the state graph \(s_t=(\mathcal{G}_t,\mathbf{H}_t)\) with adjacency matrix \(\mathbf{A}_t\), we adopt a Graph Convolutional Network (GCN)~\cite{kipf2017semi} to propagate consistency signals among references. With self-loops \(\hat{\mathbf{A}}_t=\mathbf{A}_t+\mathbf{I}\) and degree matrix \(\hat{\mathbf{D}}_t\), the \(l\)-th layer updates node features as
\begin{equation}
\mathbf{H}_t^{(l+1)}=
\sigma\!\left(
\hat{\mathbf{D}}_t^{-\frac{1}{2}}
\hat{\mathbf{A}}_t
\hat{\mathbf{D}}_t^{-\frac{1}{2}}
\mathbf{H}_t^{(l)}\mathbf{W}^{(l)}
\right),
\quad \mathbf{H}_t^{(0)}=\mathbf{H}_t,
\end{equation}
where \(\sigma(\cdot)\) denotes a nonlinear activation.
After \(L\) layers (\(L=2\)), we parameterize the action distribution by combining node-level deletion scores with graph-level operation scores:
\begin{equation}
\pi_\theta(a_t\mid s_t)=\operatorname{Softmax}\!\Big(
\Big[
\{\operatorname{MLP}_{\text{local}}(h_x^{(L)})\}_{x\in\mathcal{X}_t},\;
\operatorname{MLP}_{\text{global}}(\operatorname{Pool}(\mathbf{H}_t^{(L)}))
\Big]
\Big)_{a_t},
\end{equation}
where \(\operatorname{MLP}_{\text{local}}\) and \(\operatorname{MLP}_{\text{global}}\) are two MLP heads and \(\operatorname{Pool}\) is mean pooling.

\noindent \textbf{\textit{GRPO Optimization.}}
We optimize \(\pi_\theta\) with Group Relative Policy Optimization (GRPO)~\cite{guo2025deepseek}, which eliminates the need for a learned critic by using group statistics as a dynamic baseline. For each query \(q\), we sample a group of trajectories \(\mathcal{T}_G=\{\tau_1,\ldots,\tau_G\}\) from \(\pi_{\theta_{\text{old}}}\), compute their returns \(\{R(\tau_i)\}_{i=1}^G\), and form the normalized advantage
\begin{equation}
\hat{A}_i=
\frac{R(\tau_i)-\operatorname{Mean}(\{R(\tau_j)\}_{j=1}^G)}
{\operatorname{Std}(\{R(\tau_j)\}_{j=1}^G)+\epsilon}.
\end{equation}
We then update the policy by maximizing
\begin{equation}
\mathcal{J}(\theta)=
\mathbb{E}_{q}\left[
\frac{1}{G}\sum_{i=1}^G
\left(
\frac{\pi_{\theta}(\tau_i)}{\pi_{\theta_{\text{old}}}(\tau_i)}\hat{A}_i
-\beta\,\mathbb{D}_{\text{KL}}\!\left(\pi_{\theta}(\cdot\mid q)\,\|\,\pi_{\text{ref}}(\cdot\mid q)\right)
\right)
\right],
\end{equation}
where \(\pi_{\text{ref}}\) is a reference policy and the KL term regularizes updates.

\subsection{Evidence-Conditioned Prediction}
\label{sec:prediction}
After the agent terminates, we obtain the evolved reference set \(\mathcal{X}_T\). We then use majority voting over the labels of evolved references for diagnosis:
\begin{equation}
\hat{y}
=\argmax_{c\in\mathcal{Y}}
\sum_{x\in\mathcal{X}_T}\mathbb{I}(y_x=c).
\end{equation}

\section{Experiments}

\subsection{Experimental Setup}

\noindent\textbf{Datasets.}
Public fundus benchmarks are increasingly saturated for rare-disease diagnosis because recent retinal foundation models have been trained extensively on them~\cite{Shi2025Vision,Silva2025Foundation}, making generalization hard to assess.
We therefore evaluate on the \textit{Retina Image Bank} (RIB)~\cite{ASRSRetinaImageBank}, a public and continuously updated clinical repository that is less coupled to standard benchmarks.
From RIB, we collect 30,662 images and curate a single-label fundus dataset to reduce co-morbidity confounding.
We construct \textbf{Rare-20} (20 rare diseases; $15<N\le 75$ images/class) and \textbf{Retina-31} by adding 11 common diseases ($N>200$ images/class).
We use a 70/15/15 train/val/test split; for retrieval-based methods, the retrieval corpus is restricted to the training split to prevent test leakage.
Rare-20 contains 524/103/103 train/val/test images, and Retina-31 contains 4,737/1,001/1,023.

\noindent\textbf{Baselines and Implementation Details.}  
We compare Evo-RAD against three groups of methods. 
(1) \emph{Foundation models}: general medical VLMs (BiomedCLIP~\cite{zhang2025biomedclip}, MedCLIP~\cite{Wang2022Medclip}) and retina-specific models (EyeCLIP~\cite{Shi2025Vision}, FLAIR~\cite{Silva2025Foundation}, RetiZero~\cite{Wang2025Enhancing}).
(2) \emph{PEFT methods}: CoOp~\cite{Zhou2022Learning}, 
CLIP-Adapter~\cite{Gao2021CLIP}, XCoOp~\cite{Zhan2024Xcoop}, BiomedCoOp~\cite{Koleilat2025Biomedcoop}, TDA~\cite{Zhao2024TDA}, Tip-Adapter~\cite{Zhang2022Tip}, and DPC~\cite{Li2025DPC}. 
(3) \emph{Retrieval-based methods}: Static Retrieval and RAC~\cite{Long2023Retrieval}.
We implement Evo-RAD atop RetiZero~\cite{Wang2025Enhancing} with a candidate buffer $|\mathcal{B}|=100$ and initial retrieval size $K=8$. The policy allows at most 10 actions \(min\_size{=}2\) and is trained via GRPO with $G=8$. We report 3-seed averages using an NVIDIA RTX 4090 (24GB).

% To implement Evo-RAD, we adopt {RetiZero}~\cite{Wang2025Enhancing} as the foundation model for retrieval and adaptation baselines. Evo-RAD maintains a candidate buffer of \(N{=}100\), with an initial retrieval size \(K{=}8\), runs for at most 10 actions, and enforces a deletion minimal-size constraint \(min\_size{=}2\). During training, GRPO samples $G=8$ trajectories per query. Results are averaged over three random seeds. All experiments are conducted on a NVIDIA RTX 4090 (24GB) GPU.

\begin{table}[t]
\centering
\caption{Comparison with foundation models under zero-shot and linear probing settings. Performance is reported as mean$\pm$std for linear probing evaluation.}
\label{tab:foundation_models}
  \renewcommand{\arraystretch}{0.95}
\setlength{\tabcolsep}{2pt}
\resizebox{\columnwidth}{!}{
\begin{tabular}{l|ccc|ccc}
\toprule[1.2pt]
\multirow{2}{*}{\textbf{Method}} & \multicolumn{3}{c|}{\textbf{Rare-20}} & \multicolumn{3}{c}{\textbf{Retina-31}} \\
 & \textbf{Acc[\%]} & \textbf{F1-score[\%]} & \textbf{Sensitivity[\%]} & \textbf{ACC[\%]} & \textbf{F1-score[\%]} & \textbf{Sensitivity[\%]} \\
\midrule
\multicolumn{7}{l}{\textit{\textbf{Zero-shot Evaluation:}}} \\
BiomedCLIP \cite{zhang2025biomedclip} & 2.91 & 0.29 & 3.75 & 0.39 & 0.10 & 2.49 \\
MedCLIP \cite{Wang2022Medclip}       & 4.85 & 3.77 & 7.08 & 1.08 & 0.65 & 3.58 \\
EyeCLIP \cite{Shi2025Vision}       & 23.95 & 10.10 & 11.50 & 21.60 & 12.79 & 13.13 \\
FLAIR \cite{Silva2025Foundation}           & 20.39 & 15.86 & 21.59 & 39.00 & 24.15 & 31.08 \\
RetiZero \cite{Wang2025Enhancing}     & 25.24 & 26.37 & 29.35 & 11.53 & 12.86 & 19.34 \\
\midrule
\multicolumn{7}{l}{\textit{\textbf{Linear Probing Evaluation:}}} \\
EyeCLIP \cite{Shi2025Vision}   & 27.64 $\pm$ 5.01 & 14.48 $\pm$ 2.10 & 18.90 $\pm$ 3.74 & 30.49 $\pm$ 1.72 & 17.28 $\pm$ 1.33 & 19.15 $\pm$ 3.96 \\
FLAIR \cite{Silva2025Foundation}       & 27.51 $\pm$ 1.12 & 8.54 $\pm$ 0.36 & 12.60 $\pm$ 0.50 & 55.16 $\pm$ 0.40 & 22.81 $\pm$ 0.22 & 24.79 $\pm$ 0.23 \\
RetiZero \cite{Wang2025Enhancing} & 26.86 $\pm$ 2.02 & 10.15 $\pm$ 0.84 & 13.05 $\pm$ 1.05 & 36.62 $\pm$ 0.31 & 13.86 $\pm$ 0.28 & 15.56 $\pm$ 0.13 \\
\midrule
\rowcolor{cyan!10} \textbf{Evo-RAD (Ours)} & \textbf{46.28 $\pm$ 0.56} & \textbf{40.99 $\pm$ 0.53} & \textbf{42.43 $\pm$ 0.21} & \textbf{65.33 $\pm$ 0.28} & \textbf{51.45 $\pm$ 0.43} & \textbf{49.53 $\pm$ 0.52} \\
\bottomrule[1.2pt]
\end{tabular}
}
%\vspace{-0.5em}
\end{table}

\begin{table}[t]
\centering
\caption{Comparison with retrieval and PEFT methods (mean$\pm$std). RetiZero is the foundation model. {Params} is number of trainable parameters on Retina-31.}
\label{tab:main_sota}
\renewcommand{\arraystretch}{0.95}
\setlength{\tabcolsep}{2pt}
\resizebox{\columnwidth}{!}{
\begin{tabular}{l|c|ccc|ccc}
\toprule[1.2pt]
\multirow{2}{*}{\textbf{METHOD}} & \multirow{2}{*}{\textbf{Params}} & \multicolumn{3}{c|}{\textbf{Rare-20}} & \multicolumn{3}{c}{\textbf{Retina-31}} \\
 & & \textbf{Acc[\%]} & \textbf{F1-score[\%]} & \textbf{Sensitivity[\%]} & \textbf{Acc[\%]} & \textbf{F1-score[\%]} & \textbf{Sensitivity[\%]} \\
\midrule
\color{gray}\textit{Zero-shot} & \color{gray}0 & \color{gray}25.24 & \color{gray}26.37  & \color{gray}29.35  & \color{gray}11.53 & \color{gray}12.86  & \color{gray}19.34  \\
\color{gray}\textit{Static Retrieval} & \color{gray}0 & \color{gray}42.72 & \color{gray}35.13 & \color{gray}36.64 & \color{gray}60.41 & \color{gray}43.93 & \color{gray}41.21 \\
\midrule
CoOp \cite{Zhou2022Learning}             & 3K & 29.13 $\pm$ 0.35 & 25.31 $\pm$ 0.13 & 27.52 $\pm$ 0.75 & 20.95 $\pm$ 1.05 & 15.34 $\pm$ 0.72 & 17.44 $\pm$ 0.41 \\
CLIP-Adapter \cite{Gao2021CLIP}        & 263K & 35.40 $\pm$ 0.30 & 23.39 $\pm$ 0.50 & 26.95 $\pm$ 0.28 & 20.70 $\pm$ 0.11 & 14.44 $\pm$ 0.31 & 18.71 $\pm$ 0.25 \\
Tip-Adapter \cite{Zhang2022Tip} & 0 & 37.86 $\pm$ 0.00 & 24.45 $\pm$ 0.00 & 27.00 $\pm$ 0.00 & 22.97 $\pm$ 0.00 & 15.57 $\pm$ 0.00 & 19.12 $\pm$ 0.00 \\
RAC \cite{Long2023Retrieval}               & 23.8K & 39.48 $\pm$ 1.48 & 35.39 $\pm$ 2.55 & 40.06 $\pm$ 2.36 & 51.81 $\pm$ 0.78 & 36.71 $\pm$ 1.37 & 45.60 $\pm$ 1.71 \\
TDA \cite{Zhao2024TDA}               & 0 & 30.10 $\pm$ 1.12 & 25.73 $\pm$ 0.13 & 30.08 $\pm$ 0.11 & 15.62 $\pm$ 0.15 & 14.33 $\pm$ 0.32 & 20.72 $\pm$ 0.52 \\
XCoOp \cite{Zhan2024Xcoop}           & 12.3K & 36.57 $\pm$ 0.97 & 31.75 $\pm$ 0.91 & 31.97 $\pm$ 1.19 & 44.28 $\pm$ 1.62 & 25.53 $\pm$ 1.25 & 25.08 $\pm$ 1.09 \\
DPC \cite{Li2025DPC}               & 24.6K & 30.74 $\pm$ 1.48 & 25.66 $\pm$ 0.74 & 27.02 $\pm$ 1.28 & 34.77 $\pm$ 0.31 & 19.79 $\pm$ 3.03 & 19.32 $\pm$ 2.39 \\
BiomedCoOp \cite{Koleilat2025Biomedcoop} & 12.3K & 42.72 $\pm$ 0.97 & 36.89 $\pm$ 1.89 & 37.05 $\pm$ 1.74 & 46.20 $\pm$ 0.65 & 26.03 $\pm$ 0.57 & 25.58 $\pm$ 0.51 \\ 
\midrule
\rowcolor{cyan!10} \textbf{Evo-RAD (Ours)} & \textbf{116.4K} & \textbf{46.28 $\pm$ 0.56} & \textbf{40.99 $\pm$ 0.53} & \textbf{42.43 $\pm$ 0.21} & \textbf{65.33 $\pm$ 0.28} & \textbf{51.45 $\pm$ 0.43} & \textbf{49.53 $\pm$ 0.52} \\
\bottomrule[1.2pt]
\end{tabular}
}
\end{table}

\subsection{Main Results}
\noindent\textbf{Comparison with Foundation Models.}
Table~\ref{tab:foundation_models} shows that Evo-RAD consistently outperforms both general medical VLMs and retina-specific foundation models. On Rare-20, Evo-RAD achieves 46.28\% ACC / 40.99\% macro-F1 / 42.43\% sensitivity, outperforming the strongest baseline (RetiZero zero-shot) by a wide margin. On Retina-31, Evo-RAD reaches 65.33\% ACC / 51.45\% macro-F1, substantially exceeding linear-probing results of FLAIR and RetiZero, indicating stronger generalization across both rare and common diseases.
% Table~\ref{tab:foundation_models} shows that Evo-RAD consistently outperforms both general medical VLMs and retina-specific foundation models. On Rare-20, Evo-RAD achieves \(46.28\%\) ACC / \(40.99\%\) macro-F1 / \(42.43\%\) sensitivity, surpassing the strongest baseline (RetiZero zero-shot) by large margins. On Retina-31, Evo-RAD reaches \(65.33\%\) ACC and \(51.45\%\) macro-F1, substantially improving over linear-probing results of FLAIR and RetiZero, indicating better generalization on both common and rare diseases.

\noindent\textbf{Comparison with Retrieval and PEFT Methods.}
As reported in Table~\ref{tab:main_sota}, Evo-RAD improves over retrieval, and PEFT approaches built on the same foundation model. Notably, compared with \textit{Static Retrieval}, Evo-RAD yields consistent gains on both Rare-20 and Retina-31, and it also outperforms RAC and prompt tuning baselines, demonstrating the advantage of the self-evolving agentic retrieval framework over fixed retrieval or lightweight prompt tuning.

\noindent\textbf{Efficiency Analysis.}
Evo-RAD contains only \(116.4\)K trainable parameters (Table~\ref{tab:main_sota}), yet achieves substantially better performance. This indicates the gains mainly come from evolving retrieval rather than heavy fine-tuning, making Evo-RAD efficient to train and deploy.

\begin{figure}[!t]
\centering

\begin{minipage}[t]{0.64\linewidth}
  %\vspace{0pt} % <- remove extra top space
  \centering
  \footnotesize
  \renewcommand{\arraystretch}{0.9}
  \setlength{\tabcolsep}{4pt}
  \captionsetup{type=table}
  \caption{Ablation study on state representation, step-wise rewards, and terminal rewards.}
  \label{tab:ablation}
  \resizebox{\linewidth}{!}{%
    \begin{tabular}{lccc}
      \toprule[1.2pt]
      \textbf{Setting} & \textbf{Acc [\%]} & \textbf{F1-score [\%]} & \textbf{Sensitivity [\%]} \\
      \midrule
      \rowcolor{cyan!10} \textbf{Full (all components)} & \textbf{46.28 $\pm$ 0.56} & \textbf{40.99 $\pm$ 0.53} & \textbf{42.43 $\pm$ 0.21} \\
      \midrule
      \multicolumn{4}{l}{\textit{\textbf{(a) Ablation on State Representation:}}} \\
      $-$ \textit{Mean deviations}  & 42.74 $\pm$ 0.42 & 35.46 $\pm$ 0.54 & 36.21 $\pm$ 0.63 \\
      $-$ \textit{Base metrics}  & 44.56 $\pm$ 0.43 & 36.21 $\pm$ 0.58 & 38.41 $\pm$ 0.44 \\
      \midrule
            \multicolumn{4}{l}{\textit{\textbf{(b) Ablation on Trajectory-Level Rewards:}}} \\
      $-$ $R_{\text{acc}}$     & 43.68 $\pm$ 0.63 & 35.82 $\pm$ 0.68 & 36.40 $\pm$ 0.63 \\
      $-$ $R_{\text{purity}}$  & 40.14 $\pm$ 1.23 & 33.11 $\pm$ 1.92 & 33.30 $\pm$ 0.90 \\
      $-$ $R_{\text{density}}$ & 43.89 $\pm$ 1.60 & 36.06 $\pm$ 1.28 & 36.80 $\pm$ 0.94 \\ \midrule
    \multicolumn{4}{l}{\textit{\textbf{(c) Ablation on Step-Level Rewards:}}} \\
      $-$ $r_{\text{ins}}$ & 39.69 $\pm$ 0.84 & 34.84 $\pm$ 0.61 & 34.85 $\pm$ 0.42 \\
      $-$ $r_{\text{del}}$ & 40.61 $\pm$ 0.64 & 35.04 $\pm$ 0.54 & 35.65 $\pm$ 0.36 \\
      \bottomrule[1.2pt]
    \end{tabular}%
  }
\end{minipage}\hfill
\begin{minipage}[t]{0.34\linewidth}
  %\vspace{0pt} % <- remove extra top space
  \centering
\caption{Impact of inital retrieval size \(K\) on Rare-20.}
%\vspace{-4pt}
  \label{fig:k_analysis}
  \includegraphics[width=\linewidth]{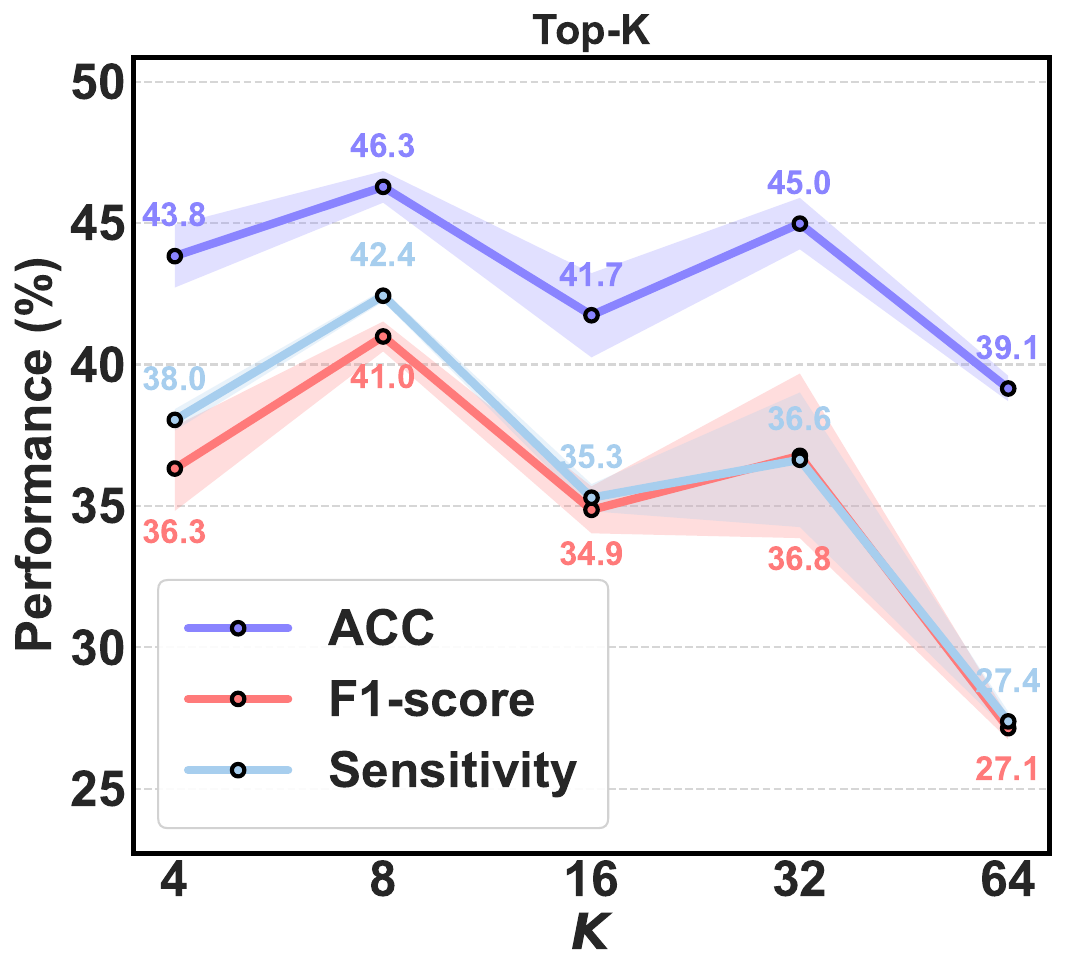}
  \captionsetup{type=figure}
\end{minipage}
\vspace{-2em}
\end{figure}

\subsection{Ablation Study}

\noindent\textbf{Ablation on State Representation.} Table~\ref{tab:ablation}(a) shows that removing either component reduces ACC. Dropping the \textit{mean deviations} (set-wise consistency) lowers ACC to 42.74\%, and removing \textit{base metrics} (individual similarity) reduces it to 44.56\%. This confirms the synergy between individual relevance and group-level consensus.

\noindent\textbf{Ablation on Trajectory-Level Rewards.}
As shown in Table~\ref{tab:ablation}(b), removing any trajectory-level reward reduces ACC: dropping \(R_{\text{acc}}\) lowers ACC to 43.68\%, while among auxiliary terms \(R_{\text{purity}}\) is most critical (40.14\% ACC without it). Removing \(R_{\text{density}}\) also degrades ACC to 43.89\%, suggesting that a well-connected reference set benefits evidence evolution.

\noindent\textbf{Ablation on Step-Level Rewards.}
As shown in Table~\ref{tab:ablation}(c), removing step-wise shaping rewards hurts ACC, dropping to 39.69\% without \(r_{\text{ins}}\) and 40.61\% without \(r_{\text{del}}\), confirming the value of dense feedback during evidence evolution.

\noindent\textbf{Ablation on Initial Reference Set Size $K$.} As shown in Fig.~\ref{fig:k_analysis}, performance peaks at \(K = 8\). Smaller \(K\) provides insufficient evidence, whereas larger \(K\) introduces more redundant/noisy references. We therefore use \(K = 8\) by default.

\section{Conclusion}
We presented Evo-RAD, a self-evolving agentic retrieval framework for rare retinal disease diagnosis that formulates evidence refinement as sequential DELETE/ INSERT/ TERMINATE decisions and learns an evidence-evolution policy via GRPO with a homogeneity-aware reward. By explicitly suppressing hub-driven distractors and consolidating clinically coherent support sets before prediction, Evo-RAD delivers consistent improvements over foundation-model baselines, standard retrieval, and PEFT variants. Future work will extend Evo-RAD to multi-label and co-morbidity settings, enabling evidence evolution under realistic mixed-pathology supervision.
% In this paper, we propose Evo-RAD, a self-evolving agentic retrieval framework that iteratively refines retrieved evidence for rare retinal disease diagnosis via DELETE/INSERT/TERMINATE actions optimized with GRPO and a homogeneity-aware reward, achieving substantial gains over foundation models, retrieval, and PEFT methods.

\noindent\textbf{Acknowledgments.} This work was partially supported by PolyU Undergraduate Research and Innovation Scheme (No. P0058439), RGC Collaborative Research Fund (No. C5055-24G), the Start-up Fund of The Hong Kong Polytechnic University (No. P0045999), the Seed Fund of the Research Institute for Smart Ageing (No. P0050946), and Tsinghua-PolyU Joint Research Initiative Fund (No. P0056509), and PolyU UGC funding (No. P0053716 and P0058848).

\bibliographystyle{splncs04}
\bibliography{ref}

@article{zhang2025biomedclip,
author = {Sheng Zhang  and Yanbo Xu  and Naoto Usuyama  and Hanwen Xu  and Jaspreet Bagga  and Robert Tinn  and Sam Preston  and Rajesh Rao  and Mu Wei  and Naveen Valluri  and Cliff Wong  and Andrea Tupini  and Yu Wang  and Matt Mazzola  and Swadheen Shukla  and Lars Liden  and Jianfeng Gao  and Angela Crabtree  and Brian Piening  and Carlo Bifulco  and Matthew P. Lungren  and Tristan Naumann  and Sheng Wang  and Hoifung Poon },
title = {A Multimodal Biomedical Foundation Model Trained from Fifteen Million Image–Text Pairs},
journal = {NEJM AI},
volume = {2},
number = {1},
pages = {AIoa2400640},
year = {2025}
}

@article{Wang2022Medclip,
  title={MedCLIP: Contrastive Learning from Unpaired Medical Images and Text},
  author={Zifeng Wang and Zhenbang Wu and Dinesh Agarwal and Jimeng Sun},
  journal={Proceedings of the Conference on Empirical Methods in Natural Language Processing. Conference on Empirical Methods in Natural Language Processing},
  year={2022},
  volume={2022},
  pages={3876-3887},
}

@article{phillips2024time,
  title={Time to diagnosis for a rare disease: managing medical uncertainty. A qualitative study},
  author={Phillips, Christine and Parkinson, Anne and Namsrai, Tergel and Chalmers, Anita and Dews, Carolyn and Gregory, Dianne and Kelly, Elaine and Lowe, Christine and Desborough, Jane},
  journal={Orphanet journal of rare diseases},
  volume={19},
  number={1},
  pages={297},
  year={2024},
}

@article{Wang2025Enhancing,
    author = {Wang, Meng and Lin, Tian and Lin, Aidi and Yu, Kai and Peng, Yuanyuan and Wang, Lianyu and Chen, Cheng and Zou, Ke and Liang, Huiyu and Chen, Man and Yao, Xue and Zhang, Meiqin and Huang, Binwei and Zheng, Chaoxin and Zhang, Peixin and Chen, Wei and Luo, Yilong and Chen, Yifan and Xia, Honghe and Fu, Huazhu},
    year = {2025},
    month = {07},
    pages={5528},
    title = {Enhancing diagnostic accuracy in rare and common fundus diseases with a knowledge-rich vision-language model},
    volume = {16},
    journal = {Nature Communications},
}

@article{Silva2025Foundation,
    title = {A Foundation Language-Image Model of the Retina (FLAIR): encoding expert knowledge in text supervision},
    journal = {Medical Image Analysis},
    volume = {99},
    pages = {103357},
    year = {2025},
    issn = {1361-8415},
    author = {Julio Silva-Rodríguez and Hadi Chakor and Riadh Kobbi and Jose Dolz and Ismail {Ben Ayed}}
}

@article{Shi2025Vision,
  author  = {Shi, Danli and Zhang, Weiyi and Yang, Jiancheng and Huang, Siyu and Chen, Xiaolan and Xu, Pusheng and Jin, Kai and Lin, Shan and Wei, Jin and Yusufu, Mayinuer and Liu, Shunming and Zhang, Qing and Ge, Zongyuan and Xu, Xun and He, Mingguang},
  title   = {A multimodal visual–language foundation model for computational ophthalmology},
  journal = {npj Digital Medicine},
  year    = {2025},
  volume  = {8},
  number  = {1},
  pages   = {381},
  month   = {6},
  issn    = {2398-6352}
}

@article{Zhou2022Learning,
    title={Learning to Prompt for Vision-Language Models},
    author={Zhou, Kaiyang and Yang, Jingkang and Loy, Chen Change and Liu, Ziwei},
    journal={International Journal of Computer Vision (IJCV)},
    year={2022}
}

@article{Gao2021CLIP,
  title={CLIP-Adapter: Better Vision-Language Models with Feature Adapters},
  author={Peng Gao and Shijie Geng and Renrui Zhang and Teli Ma and Rongyao Fang and Yongfeng Zhang and Hongsheng Li and Yu Jiao Qiao},
  journal={International Journal of Computer Vision},
  year={2021},
  volume={132},
  pages={581 - 595},
}

@inproceedings{Zhang2022Tip,
    title={Tip-adapter: Training-free adaption of clip for few-shot classification},
    author={Zhang, Renrui and Zhang, Wei and Rong, Rong and Li, Chongjian and Qiu, Zhaoshuai and Qiao, Yu and Gao, Peng},
    booktitle={European Conference on Computer Vision (ECCV)},
    pages={493--510},
    year={2022}
}

@article{luo2025llm,
  title={Llm-guided decoupled probabilistic prompt for continual learning in medical image diagnosis},
  author={Luo, Yiwen and Li, Wuyang and Chen, Cheng and Li, Xiang and Liu, Tianming and Niu, Tianye and Yuan, Yixuan},
  journal={IEEE Transactions on Medical Imaging},
  year={2025},
  publisher={IEEE}
}

@article{du2025medical,
  title={Medical Knowledge Intervention Prompt Tuning for Medical Image Classification},
  author={Du, Ye and Yu, Nanxi and Wang, Shujun},
  journal={IEEE Transactions on Medical Imaging},
  year={2025},
  publisher={IEEE}
}

@inproceedings{Long2023Retrieval,
    title={Retrieval-augmented classification for long-tail visual recognition},
    author={Long, Alexander and Yin, Wei and Ajanthan, Thalaiyasingam and Nguyen, Vu and Blair, Richard and Shen, Chunhua and van den Hengel, Anton},
    booktitle={Proceedings of the IEEE/CVF Conference on Computer Vision and Pattern Recognition (CVPR)},
    pages={6959--6969},
    year={2023}
}

@inproceedings{Zhao2024TDA,
    title = {{Efficient Test-Time Adaptation of Vision-Language Models}},
    author = {Zhao, Yifan and Zhu, Haidong and Lin, Wangli and others},
    booktitle = {Proceedings of the IEEE/CVF Conference on Computer Vision and Pattern Recognition (CVPR)},
    pages = {14201--14211},
    year = {2024}
}

@inproceedings{Zhan2024Xcoop,
    title={XCoOp: Explainable Prompt Learning for Computer-Aided Diagnosis via Concept-Guided Context Optimization},
    author={Zhan, Yuan and Wu, Chaoyi and Zhang, Ya and Wang, Yanfeng},
    booktitle={Medical Image Computing and Computer Assisted Intervention (MICCAI)},
    year={2024}
}

@inproceedings{Li2025DPC,
  title={Dpc: Dual-prompt collaboration for tuning vision-language models},
  author={Li, Haoyang and Wang, Liang and Wang, Chao and Jiang, Jing and Peng, Yan and Long, Guodong},
  booktitle={Proceedings of the Computer Vision and Pattern Recognition Conference},
  pages={25623--25632},
  year={2025}
}

@inproceedings {Koleilat2025Biomedcoop,
author = { Koleilat, Taha and Asgariandehkordi, Hojat and Rivaz, Hassan and Xiao, Yiming },
booktitle = { 2025 IEEE/CVF Conference on Computer Vision and Pattern Recognition (CVPR) },
title = {{ BiomedCoOp: Learning to Prompt for Biomedical Vision-Language Models }},
year = {2025},
pages = {14766-14776},
}

@article{Sutton1998Reinforcement,
  author={Sutton, R.S. and Barto, A.G.},
  journal={IEEE Transactions on Neural Networks}, 
  title={Reinforcement Learning: An Introduction}, 
  year={1998},
  volume={9},
  number={5},
  pages={1054-1054},
}

@misc{ASRSRetinaImageBank,
  author       = {{American Society of Retina Specialists (ASRS)}},
  title        = {Retina Image Bank},
  howpublished = {\url{https://www.asrs.org/clinical/retina-image-bank}},
  note         = {Accessed: 2026-02-25. An open-access library of nearly 30,000 downloadable retina images.},
  year         = {2026}
}

@inproceedings{kipf2017semi,
title={Semi-Supervised Classification with Graph Convolutional Networks},
author={Thomas N. Kipf and Max Welling},
booktitle={International Conference on Learning Representations},
year={2017},
}

@article{croskerry2009universal,
  title={A universal model of diagnostic reasoning},
  author={Croskerry, Pat},
  journal={Academic medicine},
  volume={84},
  number={8},
  pages={1022--1028},
  year={2009},
  publisher={Oxford University Press}
}

@article{guo2025deepseek,
  title={DeepSeek-R1 incentivizes reasoning in LLMs through reinforcement learning},
  author={Guo, Daya and Yang, Dejian and Zhang, Haowei and Song, Junxiao and Wang, Peiyi and Zhu, Qihao and Xu, Runxin and Zhang, Ruoyu and Ma, Shirong and Bi, Xiao and others},
  journal={Nature},
  volume={645},
  number={8081},
  pages={633--638},
  year={2025},
  publisher={Nature Publishing Group UK London}
}

@article{Radovanovic2010Hubs,
    title={Hubs in Space: Popular Nearest Neighbors in High-Dimensional Data},
    author={Milo{\vs} Radovanovi{\'c} and Alexandros Nanopoulos and Mirjana Ivanovi{\'c}},
    journal={J. Mach. Learn. Res.},
    year={2010},
    volume={11},
    pages={2487-2531},
}

\end{document}